\documentclass[lettersize,journal]{IEEEtran}
\usepackage{amsmath,amsfonts}
\usepackage{algorithmic}
\usepackage{array}
\usepackage[caption=false,font=normalsize,labelfont=sf,textfont=sf]{subfig}
\usepackage{textcomp}
\usepackage[most]{tcolorbox} 
\usepackage{stfloats}
\usepackage{url}
\usepackage{verbatim}
\usepackage{graphicx}
\def\BibTeX{{\rm B\kern-.05em{\sc i\kern-.025em b}\kern-.08em
    T\kern-.1667em\lower.7ex\hbox{E}\kern-.125emX}}
\usepackage{balance}

\begin{document}

\title{Shared Autonomy Assisted by Impedance-Driven Anisotropic Guidance Field}
\author{Sihan Chen$^{1,*}$, Hang Xu$^{1,\dagger}$, Yupu Lu$^{1}$, Chen Wang$^{1}$, Benfang Duan$^{2}$, Ruixing Jia$^{1}$, and Jia Pan$^{1,\dagger}$

\thanks{Manuscript accepted: March 25, 2026. ($\dagger$ Corresponding authors: Jia Pan and Hang Xu.)}

\thanks{$^{1}$Sihan Chen, Hang Xu, Yupu Lu, Chen Wang, Ruixing Jia, and Jia Pan are with the School of Computing and Data Science, The University of Hong Kong, Hong Kong, China
{\tt\footnotesize sihanchen@connect.hku.hk, xuhang\_official@outlook.com, jpan@cs.hku.hk.}}
\thanks{$^{2}$Benfang Duan is with the School of Engineering, Nanjing University of Information Science and Technology, Nanjing, China.}

}
\newcommand{\blue}[1]{\textcolor{blue}{#1}}
\newcommand{\red}[1]{\textcolor{red}{#1}}

\markboth{IEEE Robotics and Automation Letters. Preprint Version. Accepted March, 2026}
{Chen \MakeLowercase{\textit{et al.}}: Shared Autonomy Assisted by Impedance-Driven Anisotropic Guidance Field} 

\maketitle

\begin{abstract}
Shared autonomy (SA) enables robots to infer human intent and assist in its achievement. While most research focuses on improving intent inference, it overlooks whether humans can understand the robot’s intent in return. Without such mutual understanding, collaboration becomes less effective, degrading user experience and task performance. 
To address this gap, previous studies have explicitly conveyed the robot intent through additional interfaces, which remain unintuitive and limited in expressiveness. 
Inspired by impedance control, we propose Impedance-Driven Anisotropic Guidance Field Enhanced Shared Autonomy (IAGF-SA), a novel paradigm that extends SA with an embodied, physically-grounded communication channel. This channel adaptively modulates the robot’s dynamic response to human input, enabling intuitive, continuous, physically-grounded robot intent communication while naturally guiding human actions.
User studies across three scenarios and two teleoperation interfaces indicate that IAGF-SA improves task performance, human-robot agreement, and subjective experience, thus demonstrating its effectiveness in enhancing human-robot communication and collaboration.
\end{abstract}



\section{Introduction}\label{intro}

\IEEEPARstart{S}{hared} Autonomy (SA) is an established paradigm in Human-Robot Interaction (HRI) that synergistically combines robotic capabilities with human decision-making to achieve shared goals \cite{dragan2013policy, jain2018recursive}. A typical framework comprises three components \cite{jain2018recursive}: (1) \textit{Goal Inference}, which interprets the user’s intent from control inputs (e.g., joystick signals); (2) \textit{Robot Decision}, where an autonomous policy generates an assistive command; and (3) \textit{Action Blending}, which fuses the user's and robot's commands. This integration allows SA systems to mitigate human limitations in situational awareness, rationality, and motor precision, thereby reducing workload and improving task performance \cite{oh2021learning, schaff2020residual}. These benefits have led to successful applications in diverse domains such as assistive manipulation \cite{jonnavittula2024sari, zurek2021situational}.

However, a fundamental asymmetry in communication persists even in successful SA systems. While the robot continuously interprets human input, the operator has little access to the robot’s internal intent-related states \cite{oh2021learning, jonnavittula2024sari, zurek2021situational}. We define \textit{transparency} as the extent to which the operator can perceive and understand such internal information \cite{alonso2018system}. 
The lack of \textit{transparency} in SA systems arises, because the blended action, despite combining human and robot commands, is sent directly to the controller for execution, without providing a dedicated semantic channel for human interpretation. This problem is exacerbated in complex interactions involving operator hesitation or changes in intent, where operators cannot assess whether their actions are being interpreted correctly or understand how to adjust input for clarity.
This transparency limitation frequently leads to longer completion times, increased subjective workload, and lower perceived collaboration quality \cite{hoegerman2024aligning, rosen2019communicating}. 

\begin{figure}[t]
      \centering
     
      \includegraphics[width=0.85\linewidth]{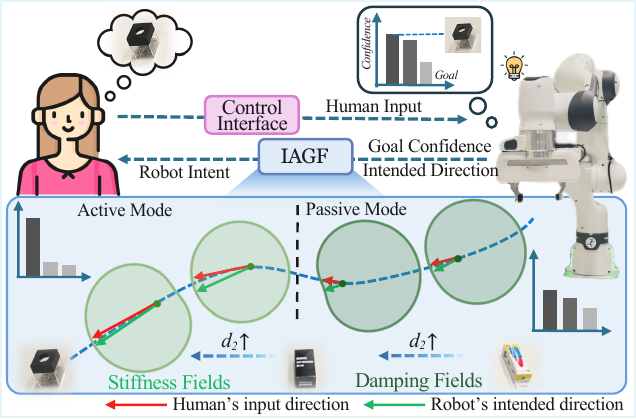}
      \vspace{-5mm}
      \caption{Illustration of human-robot bidirectional communication.}
      \label{concept}
      \vspace{-0.32in}
   \end{figure}

Enhancing robot-to-human communication is a critical pathway to improving transparency and mutual understanding in SA systems \cite{habibian2025survey}. Prior research has explored diverse modalities to convey robot intent, including visual displays \cite{hoegerman2024aligning}, light indicators \cite{backman2025novice}, and virtual reality interfaces \cite{rosen2019communicating}. However, such modalities only display the robot’s predicted intent, failing to deliver actionable operational support to users. 
While haptic shared control \cite{zhang2021haptic, coffey2022collaborative} provides action-level haptic guidance through a dedicated device, existing research is largely limited to satisfying low-level environmental constraints rather than high-level task goals communication and often enforces passive human compliance, functioning as unilateral low-level guidance rather than fostering high-level robot-to-human communication. These critical limitations underscore the need for an intuitive communication strategy that seamlessly integrates high-level intent, confidence, and guidance into a coherent, efficient interactive experience.

To address this need, impedance control \cite{bazzi2020goal, xing2021iterative} offers a promising pathway. Widely adopted in physical HRI, it models the robot's dynamics as a mass-spring-damper system, enabling compliant and seamless collaboration via adaptive damping and stiffness modulation. Our insight is that the robot’s variations in resistance or compliance can inherently communicate its intent and provide motion guidance, establishing a natural, physically-grounded embodied communication channel that eliminates the need for external interfaces \cite{hoegerman2024aligning, backman2025novice, rosen2019communicating}. Specifically, the robot can convey rich, high-bandwidth information through this embodied interaction. Building on this insight, we introduce a directionally-specific modulation mechanism, allowing the robot to encode intent through anisotropic impedance adaptation for more nuanced communication.

In this paper, we introduce \textit{Impedance-Driven Anisotropic Guidance Field Enhanced Shared Autonomy} (IAGF-SA), a novel paradigm extending standard SA with an embodied, physically-grounded robot-to-human communication channel through anisotropic impedance adaptation.
Concretely, IAGF is an anisotropic, circle-like virtual field whose radial length encodes the robot's directional preference. The preference magnitude for the human’s input direction is determined by its alignment with the robot’s intended direction.
This field acts as an impedance modulation map that scales damping or stiffness parameters by the aforementioned alignment degree, and applies them along the current motion direction to shape the robot’s dynamic response — rendering it either sluggish or agile. This enables operators to intuitively perceive directional alignment with the robot's intent and receive implicit guidance for smoother coordination.

IAGF dynamically adapts to the robot’s intent and system state through two mechanisms: adjusting the overall size and shape of the virtual field, and switching between two complementary modes (passive damping modulation and active stiffness modulation). These mechanisms enable the communication channel to convey continuous, rich information that reflects the evolution of interaction. Furthermore, multiple IAGFs can be instantiated for distinct objectives (e.g., goal-directed tasks, singularity avoidance) and fused via a superposition mechanism, ensuring broad applicability across diverse interaction scenarios.

We validated IAGF-SA on daily grasping tasks. Compared to pure teleoperation \cite{niemeyer2016telerobotics} and standard SA \cite{dragan2013policy, jain2018recursive}, IAGF-SA consistently improved task performance, human-robot alignment, and subjective experience across varied scenarios and interfaces, demonstrating its effectiveness in enhancing human-robot communication and collaboration.

\section{Related Works}

\subsection{Communication in Shared Autonomy}

Bidirectional communication \cite{habibian2025survey} is essential for SA, and prior work has explored various interfaces to convey a robot’s internal state or intention to the human. Screens are the most widely used interface \cite{hoegerman2024aligning, rossi2021evaluation, cleaver2021dynamic}, but their detachment from the robot’s workspace forces operators to split attention, which breaks spatial grounding, increases cognitive load, and impairs interaction fluency \cite{suzuki2022augmented}. 
Augmented reality (AR) and virtual reality (VR) address this spatial misalignment by embedding virtual cues directly into the the workspace \cite{rosen2019communicating, wang2023explainable}, yet they often require additional or costly hardware, limiting their feasibility in industrial or unstructured scenarios. 
Colored lights \cite{backman2025novice, song2019designing} and auditory signals \cite{tellex2020robots, unhelkar2020decision} are low-cost and intuitive, but convey only discrete, low-dimensional information, which limits their expressive capacity.
Across these modalities, a shared limitation is that they reveal the robot’s intent but provide no guidance on how the human should response, leaving communication informative but unable to meaningfully support the user’s next action.
Haptic shared control or haptic guidance \cite{zhang2021haptic, coffey2022collaborative} has emerged as a promising approach to address this gap by providing corrective action-level haptic feedback. However, such guidance is typically derived from low-level environmental constraints, without interpreting and communicating the human’s high-level intent, and often enforces passive user compliance rather than collaboration.
In contrast, our work encodes rich high-level information—including the robot’s real-time confidence and human-robot intent alignment—directly into the robot’s embodied dynamic behavior, enabling spatially grounded, intuitive communication with implicit actionable guidance.

\subsection{Impedance Control in HRI}
Impedance control is widely used in HRI, with recent work focusing on variable schemes that adjust stiffness and damping to improve safety, comfort, and task performance. Some methods globally increase damping near singularities or joint limits to enhance stability \cite{reyes2021safe}, while others modulate damping along the user’s force direction to reducing effort \cite{ficuciello2015variable, chen2024variable}. However, such directional modulation is entirely human-driven, reflecting only the operator's input rather than the robot's own intent. Task-driven strategies assign distinct impedance along Cartesian axes—reducing stiffness in uncertain directions and increasing it along confident ones—to balance human adjustment and precision \cite{parent2020variable, muhlbauer2024probabilistic}, while goal-oriented approaches lower damping toward task targets \cite{bazzi2020goal}. Yet these methods primarily optimize motion efficiency and do not convey the robot’s internal state.
In contrast, our work integrates impedance control into SA, using directional impedance as a dedicated communication channel. Coupling direction-dependent stiffness or damping to the robot’s inferred intent, our method lets the robot externalize its directional preferences, boosting decision transparency and intuitive interaction guidance.

\begin{figure*}[h]
    \centering
    \includegraphics[width=0.85\textwidth]{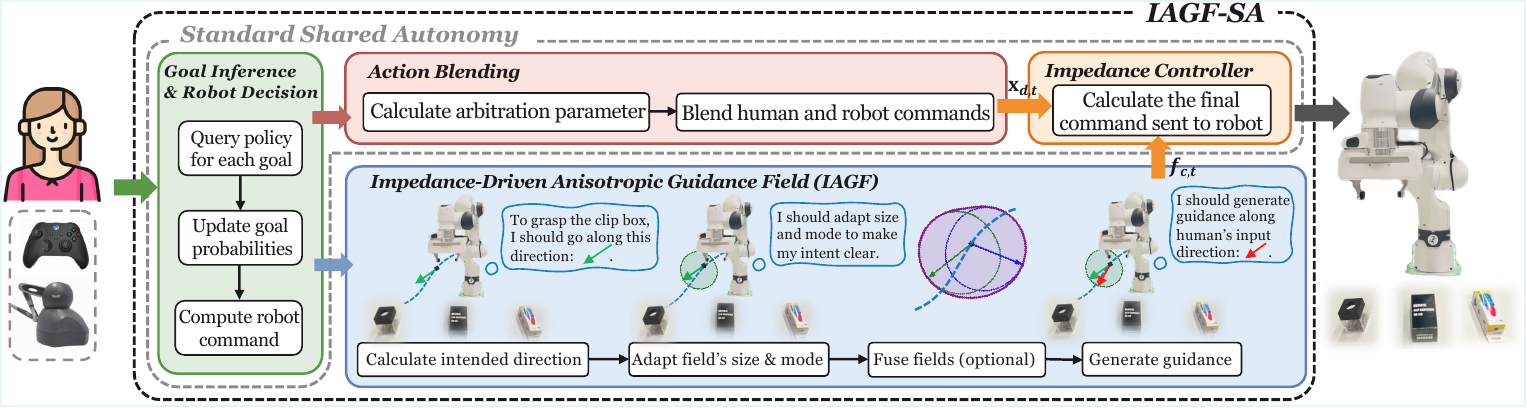}
    \vspace{-3mm}
    \caption{Overview of the IAGF-SA framework. }
    \label{frame}
    \vspace{-5mm}
\end{figure*}

\section{Overview} \label{framework}

The IAGF-SA framework enhances transparency in human–robot collaboration by improving the operator's awareness of the robot’s intent and decision-making process. It facilitates intuitive, bidirectional communication by conveying the robot’s intent inferred from human input, while also providing implicit guidance cues that promote smoother and more effective collaborative behavior. As shown in the overview in Fig. \ref{frame}, IAGF-SA extends the conventional SA framework \cite{jain2018recursive}—comprising \textit{goal inference}, \textit{robot decision}, and \textit{action blending}—through the introduction of an embodied communication channel, the IAGF. 

The IAGF communication channel operates in parallel with the blending of human command $a_h$ and robot command $a_r$. Formally, the IAGF-SA framework combines the blended action $a_{SA}$ and the IAGF-generated guidance $a_c$ through an effect-level composition: 
$ a = 
\underbrace{\big((1-\beta) \, a_h + \beta \, a_{r}\big)}_{\text{$a_{SA}$}}
\oplus \, a_c$, 
where $\oplus$ denotes effect-level composition and $\beta \in [0,1]$ is the arbitration parameter. While $a_{SA}$ interpolates between human and robot commands for correction or compensation, $a_c$ functions in a complementary manner to continuously adapt the robot's behavior, conveying its internal intent while providing adaptive guidance.

The resulting robot command is implemented through a Cartesian impedance controller:
\begin{equation} 
\ddot{\mathbf{x}}_{d,t} = M^{-1} \Big( K(\mathbf{x}_{d,t} - \mathbf{x}_t) + D(\dot{\mathbf{x}}_{d,t} - \dot{\mathbf{x}}_t) + f_{c,t} \Big),
\label{dynamics}
\end{equation}
where $\mathbf{x}_t$ and $\dot{\mathbf{x}}_t \in \mathbb{R}^m$ are the actual position and velocity of the end-effector; $\mathbf{x}_{d,t} = a_{SA}$ and $\dot{\mathbf{x}}_{d,t} \in \mathbb{R}^m$ are the desired values derived from action blending; $M, K, D \in \mathbb{R}^{m \times m}$ are the inertia, stiffness, and damping matrices, respectively; $f_{c,t} = a_c \in \mathbb{R}^m$ is the force generated by the IAGF; and $\ddot{\mathbf{x}}_{d,t}$ is the final commanded acceleration sent to the robot. The first two terms in Eq. \ref{dynamics} establish a conventional impedance behavior driving the end-effector toward the desired position, while $f_{c,t}$ (i.e., $a_c$) modulates the robot's dynamic response to achieve two primary objectives: encoding informative cues about the robot's intent and confidence, and providing natural passive or active guidance. This integrated approach ensures intuitive perception of $a_c$ without requiring operators to divert attention to external displays, thereby maintaining focus on the primary task while receiving implicit guidance through the robot's physical response.

\section{Methodology}\label{method}

This section details the IAGF. We begin by presenting its unified representation and two operating modes in Sec. \ref{method1}, which also explains how the IAGF generates the guidance force $f_{c,t}$ in Eq.~\ref{dynamics}. Building on this representation, Sec. \ref{method2} and \ref{method3} introduce two specialized IAGF instantiations tailored for enhancing task execution and avoiding robot singularities, respectively. Finally, Sec. \ref{method4} demonstrates how these IAGF variants can be integrated to seamlessly balance task progression with singularity avoidance.

\subsection{ Unified Representation}\label{method1}

For clarity, we define IAGF in 2D, though it extends naturally to higher dimensions. At each time instant $t$, the robot generates an IAGF centered at the origin of the end-effector's local Cartesian frame, operating alongside the action blending process. Given the robot's intended direction, represented by a unit vector $\mathbf{v}_r \in \mathbb{R}^2$ (with $\|\textbf{v}_r\|=1$)—later contextualized as $\mathbf{v}_{r,I}$ or $\mathbf{v}_{r,S}$—the IAGF is defined as:
\begin{equation} 
d(\phi) = d_1 \pm d_2 \, \mathbf{u}(\phi)^{\mathsf{T}} \mathbf{v}_r,
\label{gf}
\end{equation}
where $d_1$ and $d_2$ are non‑negative scalar parameters determining the radial length, and $d_2$ is always smaller than $d_1$. $\mathbf{u}(\phi) \in \mathbb{R}^2$ is a unit direction vector parameterized by the angle $\phi \in (-\pi, \pi]$, i.e., $\mathbf{u}(\phi) = [\cos \phi, \sin \phi]^{\mathsf{T}}$. This formulation yields a circle-like field where the radial length $d(\phi)$ in each direction depends on its angular alignment with $\mathbf{v}_r$. The resulting anisotropy enables the IAGF to act as an information-rich communication channel, encoding the robot’s \textit{directional preference} through varying radial lengths. A smaller angular deviation signifies a stronger preference, represented by a shorter radial length in \textit{damping field} or a longer one in \textit{stiffness field}, corresponding to the two operating modes elaborated below.

\subsubsection{Passive Mode}

When the negative sign is applied in Eq.~\ref{gf}, i.e., $d(\phi) = d_1 - d_2 \, \mathbf{u}(\phi)^{\mathsf{T}} \mathbf{v}_r$, the IAGF operates in a \textit{passive mode}. 
In this mode, the robot provides minimal, non-intrusive damping cues that act as a subtle reference rather than steering the operator in any specific direction. 
These cues enables the human to perceive and correct deviations between their input and the robot’s intent.
In this configuration, the IAGF functions as a \textit{damping field}, where the anisotropic radial lengths represent the damping magnitude along each respective direction, as illustrated in Fig.~\ref{field2}(a). A stronger robot preference—corresponding to a smaller angle between the current direction and $\mathbf{v}_r$—is indicated by a shorter radial length. Specifically, the radial length reaches its minimum $d_1 - d_2$ along the robot’s intended direction $\mathbf{v}_r$, represented by the dark green arrow in Fig.~\ref{field2}(a). This minimal length reflects the robot’s highest preference in that direction. As a result, when the human’s input aligns with $\mathbf{v}_r$, they experience the least resistance.

By substituting the unit vector of the human’s intended direction, $\mathbf{v}_h$ into Eq.~\ref{gf} in place of $\mathbf{u}(\phi)$, the corresponding radial length $d_h$ (denoted by the red arrow) is obtained. The guidance force $f_{c,t}$ is then computed as a damping-based term providing passive guidance: $f_{c,t} = K_d \, d_h \left( \dot{\mathbf{x}}_{d,t} - \dot{\mathbf{x}}_t \right)$, where $K_d$ is a constant gain mapping the radial length to the physical damping value. To ensure this term serves as a strictly passive damping term, we set $\dot{\mathbf{x}}_{d,t}$ to zero, such that the damping strength increases monotonically with $d_h$.

\subsubsection{Active Mode}

When a positive sign is applied in Eq.~\ref{gf}, i.e., $d(\phi) = d_1 + d_2 \, \mathbf{u}(\phi)^{\mathsf{T}} \mathbf{v}_r$, IAGF operates in an \textit{active mode} where the robot actively guides the human's actions toward the intended direction, thereby improving task efficiency and reducing physical and mental effort. IAGF is referred to as a \textit{stiffness field}, where the anisotropic radial lengths represent the stiffness magnitude along each respective direction, as shown in Fig.~\ref{field2}(b), with a stronger preference resulting in a longer radial length. The guidance force $f_{c,t}$ is formulated as a stiffness term providing active guidance toward $\mathbf{x}_{d,t}$: $
f_{c,t} = K_p d_h \left( \mathbf{x}_{d,t} - \mathbf{x}_t \right)$, where $K_p$ is a constant gain that maps the radial length to the actual stiffness value. 

Notably, the robot's intended direction $\mathbf{v}_r$, the two operating modes, and the scalars $d_1$ and $d_2$ vary consistently during interaction according to application-specific principles, establishing a continuous channel that reflects changes in the robot's intent transparently and intuitively.

In the following subsections, we present two instantiations of the IAGF for different purposes: enhancing execution toward the intended goal (see Sec.~\ref{method2}) and avoiding robot singularities within the task space (see Sec.~\ref{method3}).

\subsection{Intent-aware Guidance Field}\label{method2} 
The IAGF is instantiated as an Intent-aware Guidance Field (\texttt{IntGF}), designed to convey the robot's inferred human intent to enhance intent transparency and task efficiency.

Specifically, during the SA process, the human's input direction is represented by a unit vector $\mathbf{v}_h$, and the robot's intended direction is denoted as $\mathbf{v}_{r,I}$. At each time step, the robot performs goal inference (detailed in Sec.~\ref{inference}) to predict the human’s intent and obtains a corresponding confidence level $C$. 
Accordingly, the radial length $d_{h,I}$ of the guidance field along the direction of $\mathbf{v}_h$ is defined as:
\begin{equation} 
d_{h,I} = d_1 + \text{sign}(C - C_{\mathrm{th}}) \, d_2 \mathbf{v}_h^{\mathsf{T}} \mathbf{v}_{r,I} \notag \end{equation}
where $C_{\mathrm{th}}$ is a predefined confidence threshold.

\begin{figure}[b]
      \vspace{-0.20in}
      \centering
      \includegraphics[width=0.9\linewidth]{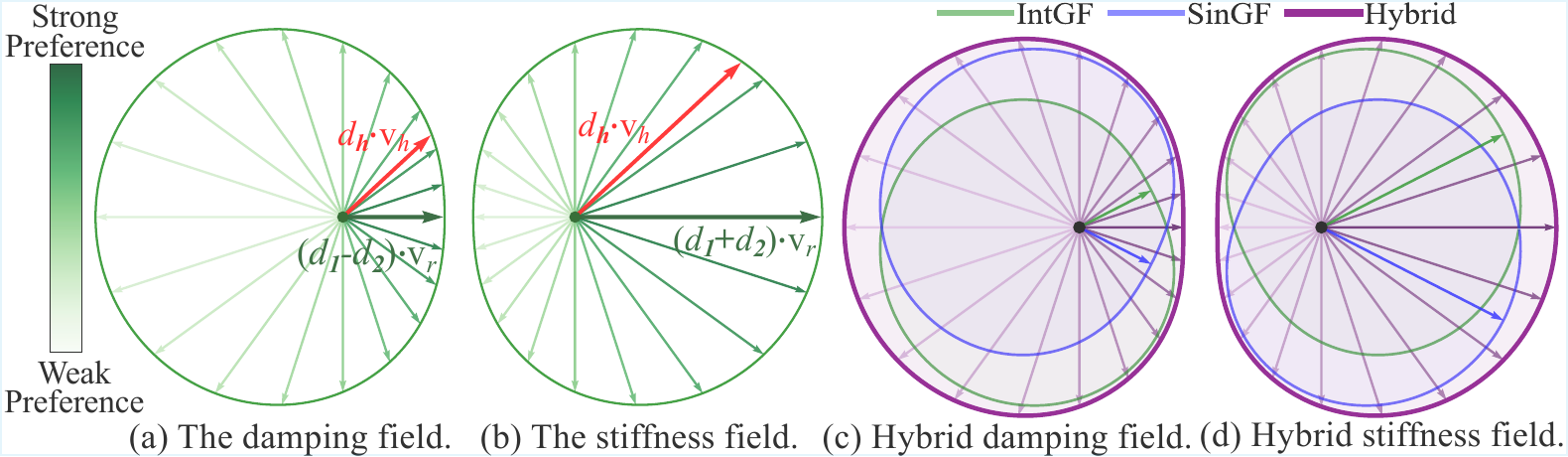} 
      \vspace{-0.1in}
      \caption{Representations of the IAGF. }
      \label{field2}  
\end{figure}

When the confidence $C$ is below the threshold $C_{\mathrm{th}}$, the robot operates with low certainty about human intent. The \texttt{IntGF} then functions in a \textit{passive} mode, referred to as the Intent-aware Damping Guidance Field (\texttt{IntGF-D}; see Fig.~\ref{field2}(a)), where damping signals deviations from the inferred intent without imposing misleading directional cues.
Conversely, when $C$ exceeds $C_{\mathrm{th}}$, the system switches to an \textit{active} mode, forming the Intent-aware Stiffness Guidance Field (\texttt{IntGF-S}; Fig.~\ref{field2}(b)), where stiffness is used to amplify the robot’s intended direction, improving intent communication and mutual understanding.  
While $d_1$ remains constant, $d_2$ varies with $C$:
\begin{equation}
d_2 = 
\begin{cases}
\dfrac{d_1}{C_{\mathrm{th}}} \, C, & C < C_{\mathrm{th}}, \\[8pt]
\dfrac{d_1}{1 - C_{\mathrm{th}}} \, C - \dfrac{d_1 C_{\mathrm{th}}}{1 - C_{\mathrm{th}}}, & C_{\mathrm{th}} \le C \le 1.
\end{cases} \notag
\end{equation}
Across both mode, $d_2$ increases with $C$.By definition, $d_2$ governs the maximum deviation from the base length  $d_1$. A larger $d_2$ widens the radial length difference in all directions, enhancing the anisotropy and clarifying the corresponding communication, as illustrated in Fig. \ref{concept}.

\subsection{Singularity-aware Guidance Field}\label{method3} 

In addition to assisting humans in task execution, robots must also address hidden risks within the workspace that may not be perceptible to the operator. Among these, singular configurations pose a significant threat, as they can result in loss of controllability and unpredictable system behavior. To mitigate this risk, we introduce a Singularity-Aware Guidance Field (\texttt{SinGF}), which communicates the robot’s intent to avoid singular configurations and intuitively guides the operator away from such regions. The robot's proximity to singularities is quantified using the manipulability measure  \cite{yoshikawa1985manipulability}, defined as: $m(\mathbf{q}) = \sqrt{\det\left(\mathbf{J}(\mathbf{q})\, \mathbf{J}(\mathbf{q})^{\mathsf{T}}\right)}$, 
where $\mathbf{q} \in \mathbb{R}^{n}$ is the joint configuration vector and $\mathbf{J}(\mathbf{q}) \in \mathbb{R}^{m \times n}$ is the manipulator Jacobian matrix at configuration $\mathbf{q}$, which maps joint velocities to end-effector velocities. To obtain a robust and effective guidance direction in Cartesian space, we compute the change in manipulability $\Delta m$ \cite{reyes2021safe} and the normalized displacement of the end-effector $\Delta \mathbf{x}$ as:
\begin{equation} 
    \begin{aligned}
    \Delta m = m(\mathbf{q}_k) - m(\mathbf{q}_{k-1}), \ \Delta \mathbf{x} &= \frac{\mathbf{x}_k - \mathbf{x}_{k-1}}{\left\| \mathbf{x}_k - \mathbf{x}_{k-1} \right\|},
    \end{aligned} \notag
\end{equation}
where $k$ and $k-1$ denote consecutive time steps. The robot’s intended direction for singularity avoidance, $\mathbf{v}_{r,S}$, is then given by $\mathbf{v}_{r,S} = \text{sign}(\Delta m)\Delta\mathbf{x} $. Accordingly, \texttt{SinGF} is:
\begin{equation}
d_{h,S} =
\begin{cases}
\text{None}, & m(\mathbf{q}) > m_{\mathrm{th}}, \\[2pt]
d_1 - d_2 \, \mathbf{v}_h^{\mathsf{T}} \mathbf{v}_{r,S}, & m_{\mathrm{crit}} < m(\mathbf{q}) \le m_{\mathrm{th}}, \\[2pt]
d_1 + d_2 \, \mathbf{v}_h^{\mathsf{T}} \mathbf{v}_{r,S}, & m(\mathbf{q}) \le m_{\mathrm{crit}},
\end{cases} \notag
\end{equation}
where $m_{\mathrm{th}}$ and $m_{\mathrm{crit}}$ are warning and critical manipulability thresholds, respectively, with $m_{\mathrm{th}} > m_{\mathrm{crit}}$. The size of the guidance field adapts dynamically with $m(\mathbf{q})$ through $d_2$:
\begin{equation}
d_2 = 
\begin{cases}
0, & m(\mathbf{q}) > m_{\mathrm{th}}, \\[2pt]
d_1 \, \dfrac{m_{\mathrm{th}} - m(\mathbf{q})}{m_{\mathrm{th}} - m_{\mathrm{crit}}}, & m_{\mathrm{crit}} \le m(\mathbf{q}) \le m_{\mathrm{th}}, \\[2pt]
d_1 \, \dfrac{m_{\mathrm{crit}} - m(\mathbf{q})}{m_{\mathrm{crit}}}, & m(\mathbf{q}) < m_{\mathrm{crit}}.
\end{cases} \notag
\end{equation}
As $m(\mathbf{q})$ decreases (indicating a trend toward singularity), $d_2$ increases, thereby amplifying the robot’s intent communication as the system approaches hidden risks. The operating principles of the \texttt{SinGF} are summarized below:
\begin{itemize}
\item $m(\mathbf{q}) > m_{\mathrm{th}}$: The manipulator operates in a well-conditioned region, and the \texttt{SinGF} remains inactive.
\item $m_{\mathrm{crit}} < m(\mathbf{q}) \le m_{\mathrm{th}}$: The manipulator approaches a singularity. The \texttt{SinGF} functions as a damping field (\texttt{SinGF-D}), operating in a passive mode that conveys risk and resists motions toward singularities.
\item $m(\mathbf{q}) \le m_{\mathrm{crit}}$: The manipulator is near a singularity. The \texttt{SinGF} acts as a stiffness field (\texttt{SinGF-S}), providing active assistance to steer away from the singular region.
\end{itemize}

\subsection{Hybrid Guidance Field}\label{method4} 
This section describes how \texttt{IntGF} and \texttt{SinGF} are fused to form a Hybrid Guidance Field, which jointly assists the human operator within the IAGF-SA framework.

\subsubsection{Homogeneous Hybrid}

When \texttt{IntGF} and \texttt{SinGF} share the same field form (i.e., both are damping fields or both are stiffness fields), they can be directly fused. Consequently, homogeneous hybrid guidance fields comprise two types:
1) \texttt{IntGF-D} and \texttt{SinGF-D}; 
2) \texttt{IntGF-S} and \texttt{SinGF-S}; 
The size of the hybrid guidance field along the human's input direction is given by $d_h = \left( d_{h,I}^{\alpha} + d_{h,S}^{\alpha} \right)^{1/\alpha}$, where $\alpha$ determines the fusion behavior between $d_{h,I}$ and $d_{h,S}$: a larger $\alpha$ amplifies the contribution of the dominant field, whereas a smaller $\alpha$ produces a balanced average. Hybrid damping and stiffness field with $\alpha=4$ are illustrated in Fig. \ref{field2}(c) and (d).

\subsubsection{Heterogeneous Hybrid}

When \texttt{IntGF} and \texttt{SinGF} adopt different field forms (i.e., one damping and one stiffness), we implement a \textit{stiffness-prioritized condition}: the stiffness field dominates, reflecting high confidence in intent or proximity to a singularity, while the damping field is suppressed. This approach supports two types of heterogeneous hybrid guidance fields:
1) \texttt{IntGF-D} and \texttt{SinGF-S};
2) \texttt{IntGF-S} and \texttt{SinGF-D}.
Under normal conditions, the hybrid fields cooperate; in critical scenarios, such as high intent confidence or near singularity, the stiffness field prevails. Thus, the hybrid field provides both singularity-related and task-related communicative guidance.

\section{Implementation Details}\label{inference}

This section details the implementation of three components integrated into our proposed IAGF-SA framework: \textit{robot decision-making}, \textit{goal inference}, and \textit{action blending}, all of which follow the standard SA pipeline.

\subsubsection{Robot Decision}

For robot assistance synthesis, we train a library of policies using the advanced imitation learning (IL) method ACT \cite{zhao2023learning}. Each policy $\pi_g: o_t \mapsto a_{g,t}$  maps observations to actions for a specific goal $g \in \mathcal{G}$, where $o_t$ consists of the current RGB image and robot state (end-effector pose and gripper state), and each action is represented in the same format as the robot state. To enhance robustness to user variability and out-of-distribution scenarios, demonstrations are collected from multiple operators under varied initial conditions. The actions generated by each policy $\pi_g$ are then utilized for goal inference. Notably, this module is not limited to IL and can be flexibly instantiated with alternative action generation methods.

\subsubsection{Goal Inference}

We employ a Recursive Bayesian framework \cite{jain2018recursive} to update the posterior belief $P_t(g)$ over all possible goals $g$ at each time step: $
P_t(g) \propto P(a_{h,t}\mid s_t, g)  P_{t-1}(g)$.
The likelihood $P(a_{h,t} \mid s_t, g)$ is defined under the principle that human input expresses intent: its similarity to the robot's actions indicates the likelihood of goal $g$. Specifically,
$P(a_{h,t} \mid s_t, g) \propto \exp \left( \gamma \, \text{sim}_{\text{enc}}(g) + (1 - \gamma) \, \text{sim}_{\text{dir}}(g) \right)$.
Here, $\text{sim}_{\text{enc}}(g)$ denotes the cosine similarity between embeddings obtained from a 3-layer MLP encoder of the most recent 6-step human commands and the robot action sequence $a_{g, t-5:t}$, capturing multi-step behavioral alignment for stable inference; $\text{sim}_{\text{dir}}(g)$ measures instantaneous directional cosine similarity, reflecting immediate action agreement for fast responsiveness to intent change. The weight $\gamma \in [0,1]$ balances the stability and responsiveness of goal inference: sole reliance on $\text{sim}_{\text{dir}}(g)$ causes high sensitivity to input noise, while sole use of $\text{sim}_{\text{enc}}(g)$ yields smooth but slow response to goal changes. The predicted goal $g^*$ is the one with the highest posterior probability, and the prediction confidence $C$ is defined as the difference between the posterior of $g^*$ and that of the next most probable goal \cite{jain2018recursive}.

\subsubsection{Action Blending}

The goal inference results are leveraged to synthesize the robot's control command through a posterior-weighted combination of all goal-directed action sequences:
$a_{r,t} = \sum_{g \in \mathcal{G}} P_t(g) \pi_g(s_t)$.
The resulting robot command $a_{r,t}$ is then blended with human command $a_{h,t}$ via an action blending mechanism, yielding the shared autonomy command $a_{SA,t} = (1 - \beta) a_{h,t} + \beta a_{r,t}$ where $\beta$ is modulated by the goal prediction confidence $C$ \cite{jain2018recursive}.

\section{User Study}\label{userstudy} 

To evaluate the proposed method, we conducted within-subject user studies across three distinct scenarios, comparing it against two baselines: pure teleoperation (NA) \cite{niemeyer2016telerobotics} and standard SA \cite{jain2018recursive}. These studies were approved by the Human Research Ethics Committee of the University of Hong Kong.

\subsection{Experimental Design}

\subsubsection{Experimental Setup}

The experimental platform consisted of a Franka Emika Panda manipulator and an Intel RealSense L515 camera. Participants teleoperated the robot using either an Xbox 360 controller or a 3D Systems Touch haptic device to verify that the proposed method is independent of the control interface.

\begin{figure}[b]
      \vspace{-0.25in}
      \centering
      \includegraphics[width=0.83\linewidth]{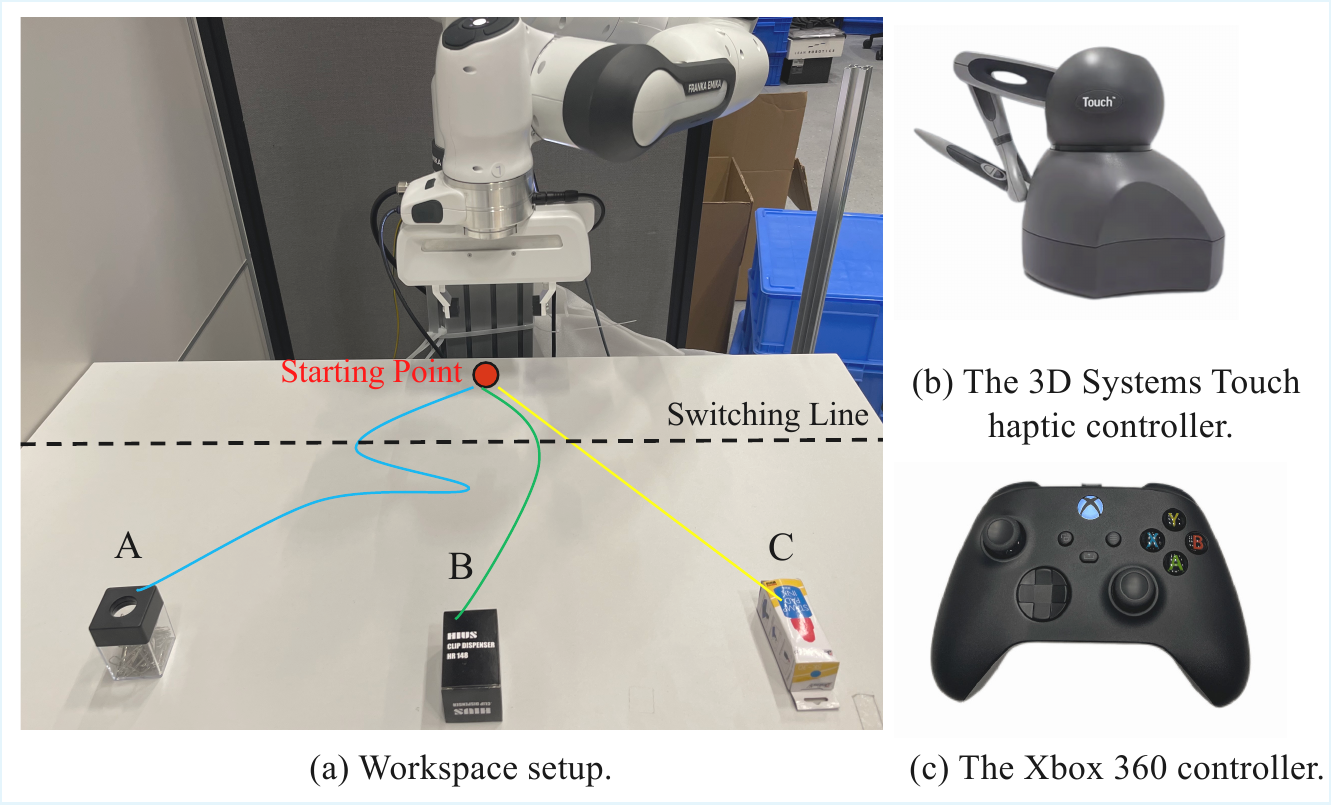}
      \vspace{-3mm}
      \caption{Experimental setup.}
      \label{scene}
\end{figure}

All three methods (NA, SA, and IAGF-SA) were implemented within an identical control framework, including the control frequency, human input processing pipeline, impedance controller (Eq.~\ref{dynamics}), and low-level controller provided by \texttt{franka-ros}. 
NA directly used raw human input, while SA and IAGF-SA utilize the blended action (Sec.~\ref{inference}) as the desired command for the impedance controller.
Notably, IAGF-SA additionally applied the $f_{c,t}$ ($K_p=80\,\mathrm{N/m}$, $K_d=10\,\mathrm{N\cdot s/m}$, $d_1=2$), whereas the $f_{c,t}$ was set to zero in NA and SA. 
Thus, these three methods differ only in action blending (absent in NA) and the application of $f_{c,t}$ (used only in IAGF-SA), ensuring a controlled comparison.

\begin{figure*}[b]
    \vspace{-0.2in}
    \centering
    \includegraphics[width=0.9\linewidth]{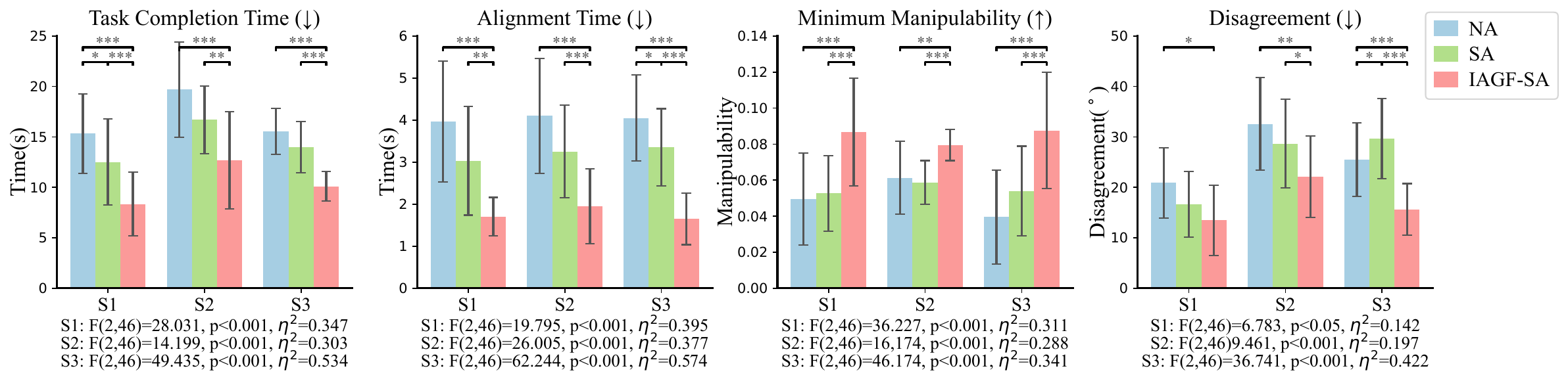}
    \vspace{-0.15in}
    \caption{Results of the objective metrics for the \textbf{\textit{joystick}} controller.}
    \label{joy_ob}
    
\end{figure*}
   
\begin{figure*}[b]
    \centering
    \vspace{-0.15in}
    \includegraphics[width=0.9\linewidth]{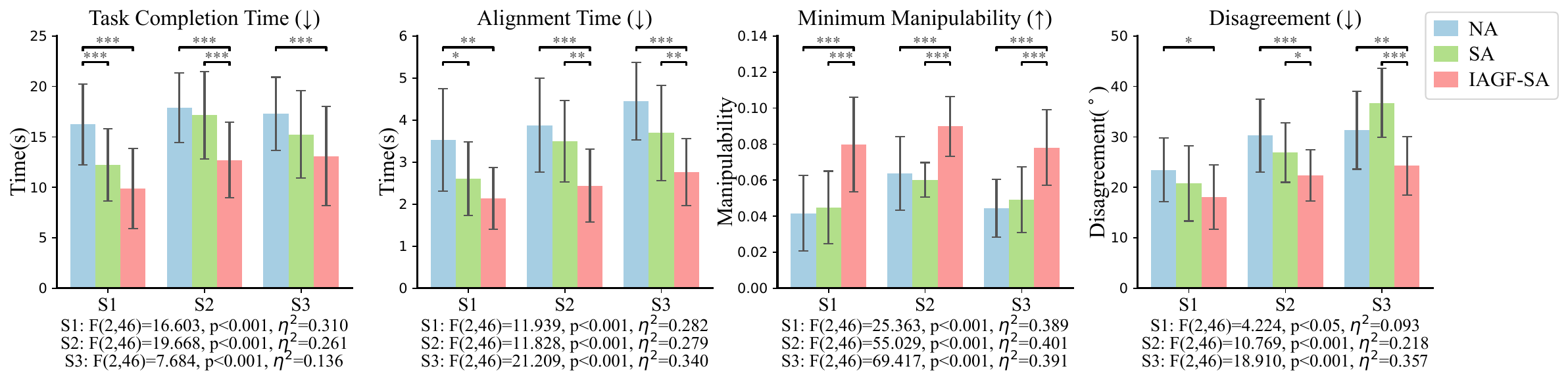}
    \vspace{-0.15in}
    \caption{Results of the objective metrics for the \textbf{\textit{haptic}} controller.}
    \label{hap_ob}
\end{figure*}

\subsubsection{Task and Scenarios} \label{exp-task-scenario}
In the user study, participants were instructed to teleoperate the robot to grasp stationery items $A$, $B$, and $C$ placed on a desk (Fig.~\ref{scene}). Unlike prior work, we emphasize the complex and dynamic nature of HRI. Human behavior can be highly variable due to changing intent, unskilled actions, or irrational decisions; users may persist with a single goal, frequently switch goals, or issue task-irrelevant commands, reducing efficiency. To assess whether the proposed guidance field addresses these challenges by providing assistance and communication, we designed three distinct scenarios:
\begin{itemize}
    \item \textit{$\textbf{S1}$ - Single-minded Scenario}:  The user selects one goal at the start and maintains it throughout the task. This scenario includes three tasks labeled by their goals: $A$, $B$, and $C$ (yellow trajectory in Fig.~\ref{scene}).
    \item \textit{$\textbf{S2}$ - Goal-switching Scenario}:  The user changes goals during execution, including transitions such as $A \rightarrow C$, $C \rightarrow A$, $B \rightarrow A$, $B \rightarrow C$ (green trajectory in Fig.~\ref{scene}).
    \item \textit{$\textbf{S3}$ - Indecisive-exploration Scenario}: The user exhibits hesitation and repeatedly alternates between goals, due to lack of focus or deliberate consideration. This scenario includes four tasks: $A \rightarrow B \rightarrow A$, $C \rightarrow B \rightarrow C$, $B \rightarrow A \rightarrow B$, $B \rightarrow C \rightarrow B$ (blue trajectory in Fig.~\ref{scene}).
\end{itemize}

These scenarios were designed to reflect increasing levels of interaction complexity in human-robot collaboration, covering both ideal and challenging cases. While $S1$ represents smooth, goal-aligned cooperation, $S2$ and $S3$ capture situations where changes in human intent cause the robot's predictions to fluctuate between certain and uncertain states. Such instability complicates consistent assistance and heightens the need for the human to understand the robot's internal intent to prevent misunderstandings and performance degradation. Our proposed method directly addresses this challenge, and these scenarios were selected to demonstrate its advantages across varying interaction complexities.

To standardize trajectories in Scenarios 2 and 3, which involve goal switching, we introduced a mechanism to reduce variability. A virtual switching line (invisible to the operator) was defined in the workspace, as shown in Fig.~\ref{scene}. When the robot reached this line, participants received an audio prompt instructing them to switch goals.

\subsubsection{Independent Variables}

Each task was performed by participants under three collaboration methods:

\begin{itemize}
        \item \textit{Pure Teleoperation} \textit{\textbf{(}No Assistance: \textbf{NA)}}: The robot executes the operator’s teleoperation commands without providing assistance.
    \item \textit{Shared Autonomy} \textbf{\textit{(SA)}}: The robot assists the operator through action blending.
    \item \textit{Communicative Shared Autonomy} \textbf{\textit{(IAGF-SA)}}: The robot provides both action blending and embodied communicative assistance generated by our method.
\end{itemize}

\subsubsection{Dependent Measures} \label{exp-measure}

To evaluate the performance of the proposed method, we used two categories of dependent measures: objective and subjective.

\textbf{Objective measures} included the following four metrics for quantitative evaluation of task performance and robotic assistance characteristics:
\begin{itemize}
\item \textit{Task Completion Time}: Time elapsed from task initiation to successful object grasp.
\item \textit{Disagreement}: One minus the mean cosine similarity between human and robot commands, reflecting directional inconsistency.
\item \textit{Alignment Time}: Time spent fine-tuning the end-effector near the target for accurate grasping.
\item \textit{Minimum Manipulability}: The lowest manipulability value recorded during task execution; lower values indicate proximity to singular configurations.
\end{itemize}

\textbf{Subjective measures} assessed the perceived usability and communicative quality of the system:
\begin{itemize}
\item \textit{Communicative Assistance Scale (CAS)} evaluates participants' perceptions of the clarity and effectiveness of the robot’s communicative behaviors.
\item \textit{System Usability Scale (SUS)} \cite{brooke2013sus} is a standardized questionnaire for measuring overall system usability.
\end{itemize}

\subsubsection{Participants and Procedure}

Twelve participants (8 males, 4 females; mean age = 26.9 years) were recruited. Nine had robotics backgrounds and three were novices. After informed consent, participants completed a 5-minute tutorial and a 10-minute practice session.
A Latin Square design was employed to counterbalance task assignment, control device order, collaboration method, and scenario sequence. Each participant performed a total of 36 trials (2 tasks × 3 scenarios × 3 methods × 2 devices). A trial was deemed complete upon successful grasp, and and control inputs and robot states were recorded for post-analysis. After finishing trials with each device, participants completed a questionnaire (Sec.\ref{exp-measure}) following a short break.

After the study, semi-structured interviews were conducted regarding collaboration method and device preferences.

\subsubsection{Hypotheses} The following hypotheses were proposed:

\textbf{H1.} IAGF-SA improves task performance (in terms of efficiency and safety) compared to both NA and SA.

\textbf{H2.} IAGF-SA increases human–robot agreement compared to NA and SA, particularly in $S2$ and $S3$. In contrast, SA is not expected to show a clear advantage over NA.

\textbf{H3.} IAGF-SA leads to a better subjective experience for the operator.

\textbf{H4.} IAGF-SA performs effectively with both joystick and haptic controllers, independent of specific feedback channels in the control interface.

\begin{figure*}[b]
      \vspace{-0.2in}
      \centering
      \includegraphics[width=0.9\linewidth]{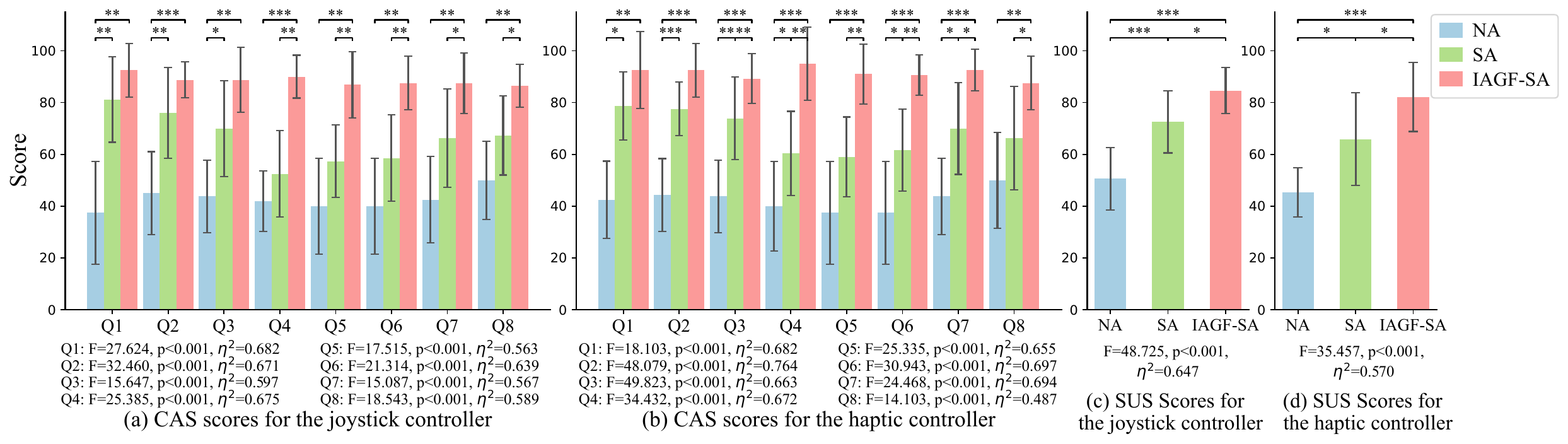}
      \vspace{-0.15in}
      \caption{Results of the CAS and SUS. }
      \label{fig_scale1}
   \end{figure*}

\subsection{Results and Analysis }

For each performance measure, a one-way repeated-measures ANOVA (rANOVA) was conducted to assess significant differences among the three collaboration methods ($p < 0.05$). When significant effects were detected, post-hoc pairwise comparisons with Bonferroni correction were applied. Significance levels are denoted as $p < 0.05$: *, $p < 0.01$: **, and $p < 0.001$: ***. Objective results are presented in Figs.~\ref{joy_ob} and~\ref{hap_ob}, and subjective results in Fig.~\ref{fig_scale1}.

\subsubsection{Effects of  Improving Task performance}
Regarding \textbf{H1}, we report three objective measures from the first three subplots of Figs.~\ref{joy_ob} and~\ref{hap_ob}.
For task efficiency, \textit{Task Completion Time} and \textit{Alignment Time} were measured. While SA reduces these times compared to NA, IAGF-SA demonstrates a further significant improvement, indicating higher efficiency than both baselines. This enhancement stems from the IntGF component (Sec.~\ref{method2}), which provides intent-aware guidance to help operators issue more effective commands. Consequently, human–robot coordination becomes smoother, unnecessary fine-tuning is reduced, and task execution is accelerated. For system safety, \textit{Minimum Manipulability} was evaluated, with smaller values indicating proximity to singularities. Supported by the SinGF component in IAGF (Sec.~\ref{method3}), IAGF-SA effectively mitigates excessive decreases in manipulability, maintaining it within a safe range and indicating lower risk of singular configurations alongside improved system stability. These findings support \textbf{H1}.

\subsubsection{Effects of  Improving Human–Robot Agreement}
Regarding \textbf{H2}, human–robot \textit{Disagreement} was measured, as shown in the last subplots of Figs.~\ref{joy_ob} and~\ref{hap_ob}.
IAGF-SA significantly reduces \textit{Disagreement} compared to SA and NA, whereas SA shows no notable improvement over NA. This effect is pronounced in complex scenarios $S2$ and $S3$, where goal changes occur during task execution. In these cases, SA fails to reduce disagreement and even underperforms NA in $S3$, suggesting that assistance without effective communication may exacerbate misunderstandings in complex interactions, leading to operator confusion and increased disagreement.
In contrast, IAGF-SA leverages IAGF to explicitly convey the robot's intent information, enhancing the human-robot understanding and alignment. The reduction in disagreement subsequently contributes to improved task performance. These results support \textbf{H2}.

\subsubsection{Effects of Improving Use Experience}

Regarding \textbf{H3}, two subjective measures were employed: the standardized \textit{SUS} and a custom \textit{CAS}. The CAS assessed participants’ perceptions of the robot’s guidance and communication behaviors. Participants rated each collaboration method on a 0–100 scale based on the following items:

\noindent\textit{Q1.} The robot helped me during the task.  

\noindent\textit{Q2.} The robot helped me complete the task more effectively.  

\noindent\textit{Q3.} The robot was able to understand my intentions.

\noindent\textit{Q4.} I was able to perceive the robot’s cues.  

\noindent\textit{Q5.} I was able to understand the robot’s cues.  

\noindent\textit{Q6.} I adjusted my operation based on the robot’s cues. 

\noindent\textit{Q7.} I trusted the system to provide appropriate assistance.  

\noindent\textit{Q8.} Please rate your overall collaboration.

CAS results (Figs.~\ref{fig_scale1}(a) (b)) show IAGF-SA consistently outperformed SA and NA, effectively conveying robot intent (Q4) and providing meaningful guidance (Q6). This led to enhanced mutual understanding and trust (Q3, Q5, Q7), and higher overall collaboration ratings (Q8). IAGF-SA also achieved highest SUS scores (Figs.~\ref{fig_scale1}(c)(d)), confirming significantly improved usability. These findings support \textbf{H3}.

\subsubsection{Interface-Independent Performance}

For \textbf{H4}, quantitative measures were not directly compared between control interfaces due to confounding factors from distinct implementations and user experiences. However, consistent superiority of IAGF-SA over SA and NA across both interfaces (as supported by \textbf{H1–H3}) provides indirect evidence that its communicative advantage stems from the robot’s intrinsic response behavior, rather than interface-specific feedback.

Post-study interviews revealed a dominant preference for IAGF-SA (10 participants) over NA and SA (1 each). Device preference was mixed: 7 favored the haptic interface for its physical feedback, while 5 preferred the joystick for its simplicity and gaming familiarity. Notably, all 3 novices preferred IAGF-SA with haptic controller for its intuitive guidance.

\section{Conclusion}

This work introduced IAGF-SA, a novel shared autonomy framework that integrates an embodied robot-to-human communication channel. By adaptively shaping the robot's dynamic response, this channel continuously conveys the robot's intent and provides actionable guidance, fostering decision quality, partner alignment, and user engagement for smoother human-robot teamwork. User studies confirmed significant improvements in task performance, human-robot agreement, and subjective experience across diverse scenarios and teleoperation interfaces, underscoring that intent transparency is essential for effective collaboration.

Several directions remain for future work. First, while IAGF-SA is broadly applicable, it holds particular advantages in scenarios with ambiguous intent, multiple feasible goals, or safety-critical requirements (e.g., collaborative assembly or precise manipulation in cluttered environments), where conveying the robot's internal state helps mitigate intent misalignment and enables safer interaction. Second, extending IAGF from 2D to higher-dimensional settings is a natural progression; its unified structure also suggests potential for broader HRI applications beyond task execution and singularity avoidance. Third, post-study interviews indicated that prior user experience may influence system preferences, suggesting future designs should account for diverse user backgrounds. Finally, physiological signals like eye-tracking data could serve as sensitive measures of subjective experience and can be incorporated in future evaluations.









\bibliographystyle{IEEEtran}
\bibliography{main}

\end{document}